\documentclass[10pt, journal, web]{article}
\usepackage{times}
\usepackage{caption}
\usepackage[utf8]{inputenc}
\usepackage{utfsym}
\usepackage{multirow}
\usepackage[normalem]{ulem}
\usepackage{amsmath,amssymb,amsfonts}
\usepackage{algorithmic}
\useunder{\uline}{\ul}{}
\usepackage{url}
\def\BibTeX{{\rm B\kern-.05em{\sc i\kern-.025em b}\kern-.08em
    T\kern-.1667em\lower.7ex\hbox{E}\kern-.125emX}}

\usepackage{graphicx}
\usepackage{authblk}

\usepackage{longtable}
\usepackage{pdflscape}
\usepackage[T1]{fontenc}
\usepackage{subfigure}
\usepackage{bbm}
\usepackage[ruled,vlined]{algorithm2e}
\usepackage{pifont}
\usepackage{xcolor}
\usepackage{hyperref}
\usepackage{booktabs}
\hypersetup{
   colorlinks=true,
    linkcolor=blue,
   citecolor=blue,
    urlcolor=blue
}
\usepackage{makecell}
\usepackage{tabularx}
\usepackage{enumitem}

\makeatletter
\renewcommand\AB@affilsepx{, \protect\Affilfont}
\makeatother
\usepackage{tabularx}
\providecommand{\keywords}[1]
{
  \small	
  \textbf{\textit{Keywords---}} #1
}

\begin{document}

\title{\textbf{ Integrating ConvNeXt and Vision Transformers for Enhancing Facial Age Estimation}}
\author[1]{G. Maroun}
\author[1]{S.E. Bekhouche}
\author[1, 2]{F. Dornaika\thanks{Corresponding author}}
\affil[1]{\textit{University of the Basque Country}}
\affil[2]{\textit{IKERBASQUE}}

\affil[ ]{

\small\texttt{gmaroun001@ikasle.ehu.eus,  sbekhouche001@ikasle.ehu.eus, fadi.dornaika@ehu.eus}}
\date{}
\maketitle
\begin{abstract}
Age estimation from facial images is a complex and multifaceted challenge in computer vision. In this study, we present a novel hybrid architecture that combines ConvNeXt, a state-of-the-art advancement of convolutional neural networks (CNNs), with Vision Transformers (ViT). While each model independently delivers excellent performance on a variety of tasks, their integration leverages the complementary strengths of the CNNs’ localized feature extraction capabilities and the Transformers’ global attention mechanisms.Our proposed ConvNeXt-ViT hybrid solution was thoroughly evaluated on benchmark age estimation datasets, including MORPH II, CACD, and AFAD, and achieved superior performance in terms of mean absolute error (MAE). To address computational constraints, we leverage pre-trained models and systematically explore different configurations, using linear layers and advanced regularization techniques to optimize the architecture. Comprehensive ablation studies highlight the critical role of individual components and training strategies, and in particular emphasize the importance of adapted attention mechanisms within the CNN framework to improve the model's focus on age-relevant facial features. The results show that the ConvNeXt-ViT hybrid not only outperforms traditional methods, but also provides a robust foundation for future advances in age estimation and related visual tasks. This work underscores the transformative potential of hybrid architectures and represents a promising direction for the seamless integration of CNNs and transformers to address complex computer vision challenges.

\end{abstract}


 \hspace{10pt}
%%%%%%%%%%%%%%%%%%%%%%%%%%%%%%%%%%%%%%%%%%%%%%%%%%%%%%%%%%%%%%%%%%%%% 

%-------------------------------------------------------------------------

\section{Introduction}
\label{sec:introduction}
The human face is a rich source of personal attributes and serves as a critical identifier in biometric systems. Beyond identity, facial images convey information such as gender, ethnicity, emotional state, and, notably, age. Age estimation—the task of approximating an individual's age based on their facial image—has diverse real-world applications in security, marketing, and human-computer interaction. Despite its utility, age estimation is inherently challenging, especially when reliant on subjective self-reporting. Accurate age prediction also provides key insights into health, cognitive abilities, and potential lifespan \cite{bekhouche2017pyramid, Deng_2021_CVPR, ordcon2025, cilf2024, Qiao2025sagn}.

In recent years, deep learning approaches, particularly Convolutional Neural Networks (CNNs), have become the leading method for age estimation \cite{osman2018computational}. However, the advent of Vision Transformers (ViTs) has opened new possibilities for improving accuracy and performance in this domain. Initially introduced in natural language processing \cite{vaswani2017attention}, the Transformer architecture was adapted for image recognition by Dosovitskiy et al. \cite{dosovitskiy2020image}. Unlike traditional CNNs, ViTs leverage self-attention mechanisms to analyze relationships between fixed-size image patches, enabling nuanced and abstract feature interactions \cite{dosovitskiy2020image, touvron2021going}. This approach has demonstrated superior performance in various visual tasks, ranging from medical imaging \cite{chen2021vit} to remote sensing \cite{zhang2021trs}, highlighting its versatility.

However, ViTs face notable challenges, including high computational costs and memory demands, which limit their scalability to larger datasets and complex tasks. To address these limitations, hybrid models such as LeViT \cite{graham2021levit} have been developed, integrating convolutional layers into the Transformer architecture to enhance computational efficiency. While ViTs represent a paradigm shift in computer vision, ongoing research aims to improve their efficiency and broaden their applicability.

Despite their promise, vanilla ViTs exhibit certain limitations, such as a reliance on large-scale datasets and minimal local inductive bias. ConvNeXt \cite{liu2022convnet} addresses these issues by combining the global receptive fields of ViTs with the structured design of CNNs. By employing depth-wise separable convolutions, ConvNeXt captures spatial hierarchies and local spatial information more effectively, enhancing data efficiency.

Despite their success in various domains, ViTs and ConvNeXt have been underexplored in age estimation—a task requiring fine-grained analysis of facial features and subtle temporal changes. In this work, we propose a novel hybrid model that combines the local pattern recognition strengths of ConvNeXt with the global information processing capabilities of ViTs for age estimation. Our architecture comprises a ConvNeXt backbone to encode facial images into feature maps, which are subsequently reshaped and processed by a Transformer. A multilayer perceptron (MLP) head then interprets the Transformer’s output to predict the final age estimate. Additionally, we employ data augmentation to enhance training data diversity.

We evaluate our model on prominent age estimation benchmarks, including AFAD, CACD, and MORPH II, with the Transformer pre-trained on ImageNet. Our results demonstrate competitive performance, achieving a mean absolute error (MAE) of 2.26 years on MORPH II (Table \ref{tab:sota_morph2}), 4.35 years on CACD (Table \ref{tab:sota_cacd}), 4.2years on IMDB-Clean (\autoref{tab:sota_imdb_mae}) and 3.09 years on AFAD (Table \ref{tab:sota_afad}). Ablation studies further validate the efficacy of each model component and training strategy.

This study offers a novel perspective on leveraging Vision Transformers for age estimation, showcasing the potential of hybrid architectures to enhance accuracy and robustness. Our findings suggest broader applicability of this approach, paving the way for advancements in related domains and complex vision tasks. \\

\label{sec:paper_contribution}

In this work, while we do not introduce a new domain to age estimation, a well-established and extensively researched field, we bring a new perspective by incorporating the innovative Vision Transformer (ViT)\cite{surveyViT} paradigm and comparing its performance with traditional Convolutional Neural Network (CNN) models. Our contributions can be summarized as follows:

\begin{itemize}
    \renewcommand\labelitemi{-}

 \item Introduction of a novel hybrid approach that combines the robust capabilities of the state-of-the-art ConvNeXt CNN architecture with the attention mechanisms of transformers, achieving state-of-the-art results and exceptional accuracy on various datasets.
    \item Addressing a significant gap in the literature by exploring the application of transformers, which have not been extensively researched in the context of age estimation, particularly utilizing their attentional mechanisms to identify informative features in facial photographs amidst different noise levels and augmentation techniques.
    \item Development and training of both ConvNeXt and ViT models, where the strengths of ConvNeXt are synergized with the unique properties of transformers, enhancing the model's ability to recognize age-related patterns and achieve higher accuracy.
    \item Extensive experimentation and evaluation with multiple datasets, demonstrating the effectiveness and innovation of the proposed hybrid model, making a substantial contribution to the evolving landscape of age estimation methods.

\end{itemize}

%-------------------------------------------------------------------------

\section{Related work}
\label{sec:review}
% \section{related work} age estimation and transformers for age estimation
%\subsection{Related Papers}

\subsection{The famous \textbf{Convolutional Neural Networks CNN}}
\label{sec:cnn}
Convolutional Neural Networks (CNNs) have transformed computer vision by enabling models to achieve remarkable accuracy in recognizing objects within images \cite{lecun1998gradient}\cite{fukushima1980neocognitron}. The core architecture of a CNN comprises multiple convolutional layers, pooling layers, and fully connected (linear) layers. Convolutional layers utilize a set of learnable filters to extract features from input images, sliding across the image in a process called convolution to produce output feature maps. These feature maps are passed through activation functions, introducing nonlinearity to enhance the network's representational power.

Pooling layers, such as max pooling or average pooling, are used to downsample feature maps, reducing data dimensionality while retaining essential information. Fully connected layers then map the high-level features from the convolutional and pooling layers to a final classification decision \cite{krizhevsky2012imagenet}. Dropout regularization is commonly employed to mitigate overfitting during training.

CNNs have demonstrated exceptional performance in a range of computer vision tasks, including image classification, object detection, and semantic segmentation \cite{simonyan2014very}. Models such as VGG-Face \cite{parkhi2015deep} and ResNet \cite{he2016deep} have gained widespread adoption in face recognition and have achieved state-of-the-art results on age estimation datasets. These successes have spurred ongoing research, leading to progressively improved age estimation techniques.

A particularly promising development in CNN-based age estimation is cross-dataset learning \cite{zhang2022cross}, which involves training a CNN on multiple age estimation datasets simultaneously. This approach enhances the robustness and generalizability of models, allowing them to perform well on unseen data. These advancements underline the evolving capabilities of CNNs in addressing challenging tasks such as age estimation and highlight the potential for further innovation in this domain.

\subsection{\textbf{Transformers} and the \textbf{Attention} mechanism}
\label{sec:transformers}

Transformers, a groundbreaking innovation in deep learning, were introduced in 2017 with the seminal paper "Attention Is All You Need" \cite{vaswani2017attention}. Originally designed for machine translation, transformers revolutionized natural language processing (NLP) by introducing the attention mechanism. This mechanism assigns dynamic weights to words based on their relevance and relationships within a sentence, enabling more accurate and context-aware translations.

The transformative potential of transformers extended beyond NLP with the pivotal work "An Image is Worth 16x16 Words: Transformers for Image Recognition at Scale" \cite{dosovitskiy2020image}. This study adapted the attention mechanism to vision tasks by allocating attention to image patches, allowing the model to understand an image’s global context and achieve significant improvements in image recognition accuracy.

In the context of age estimation, vision transformers (ViTs) offer a promising yet underexplored approach. Age estimation requires fine-grained analysis of facial features and subtle temporal changes, making it a challenging task. By leveraging the attention and contextual understanding capabilities of transformers, ViTs present a novel perspective for tackling these challenges effectively.

A key advantage of ViTs lies in their ability to model intricate and abstract relationships among image features. Unlike convolutional neural networks (CNNs), which rely on fixed convolutional operations, ViTs use self-attention mechanisms to divide an image into patches and compute inter-patch relationships. This approach enables a more holistic representation of features, improving the accuracy of tasks like age estimation.

Recent studies have highlighted the effectiveness of ViTs in image recognition. Liu et al. \cite{li2021continuous} and Xu et al. \cite{xu2021multistageage} have demonstrated that ViTs achieve superior accuracy compared to traditional CNNs in age estimation tasks. Additionally, Touvron et al. \cite{touvron2021efficient} introduced efficient variants of ViTs that not only achieved state-of-the-art results on benchmark datasets like ImageNet but also addressed computational and memory constraints, making these models more practical for deployment.

This work builds on the foundational ideas introduced by Vaswani et al\cite{vaswani2017attention}. and their subsequent adaptation for vision tasks by Dosovitskiy et al.\cite{dosovitskiy2020image}. These seminal studies provide the conceptual framework for understanding the power and potential of ViTs in diverse applications, including age estimation.

The application of vision transformers to age estimation holds immense potential for advancing the field. By harnessing their attention mechanisms and contextual analysis capabilities, ViTs pave the way for more accurate, robust, and innovative age estimation systems. This opens new avenues for research, offering transformative possibilities for computer vision tasks that require fine-grained feature interpretation.

\subsection{Facial Age Estimation}
\label{sec:facial_age_estim}
Face-based age estimation is a challenging yet fascinating task that humans perform effortlessly in their daily lives. Despite the complexities introduced by diverse facial aging patterns, ethnic variations, and lifestyle influences, humans excel at estimating age based on facial cues. Inspired by this ability, numerous methodologies have been developed to automate age estimation, striving for enhanced accuracy and robustness. These methodologies can be categorized into key clusters based on their distinct approaches and innovations.

\textbf{Hybrid Approaches:} Hybrid methods combine complementary techniques to overcome the limitations of standalone models. Wu et al. (2018), in their work "Deep Regression Forests for Age Estimation" (DRFss \cite{shen2018deep}), integrated deep learning with regression forests, capturing complex nonlinear relationships between facial features and age while improving robustness against noise and outliers. Wan et al. (2018), in "Auxiliary Demographic Information Assisted Age Estimation with Cascaded Structure" (CasCNN \cite{wan2018auxiliary}), utilized auxiliary demographic information, such as gender and race, in a cascaded structure, achieving superior performance across diverse datasets. Similarly, Tan et al. (2017) proposed "Efficient Group-n Encoding and Decoding" (AGEn \cite{8141981}), which groups adjacent ages for more efficient learning using a deep convolutional neural network (CNN) with multiple classifiers. Wang et al. (2022) advanced hybrid approaches with "Improving Face-Based Age Estimation with Attention-Based Dynamic Patch Fusion" (ADPF \cite{wang2022improving}), introducing a Ranking-guided Multi-Head Hybrid Attention (RMHHA) mechanism to dynamically identify age-specific facial patches, addressing previous methods' limitations. More recently, the GLA-Age framework (2025) \cite{LIU2025glaage} enhanced this line of work by combining global facial context with local age-sensitive patches through multi-scale attention, demonstrating improved robustness under challenging conditions.
These hybrid models showcase how combining diverse strategies can yield more robust and accurate results, an idea central to our proposed approach.

\textbf{Innovations in Loss Functions: }
The choice of loss function significantly impacts training efficiency and model performance. Pan et al. (2018), in "Mean-Variance Loss for Deep Age Estimation from a Face" (CE-MV \cite{rothe2018deep}), introduced a loss function that penalizes both prediction errors and variance, enhancing adaptability for aesthetic assessments. Akbari et al. (2021), in "A Flatter Loss for Bias Mitigation in Cross-Dataset Facial Age Estimation" (SGD \cite{akbari2021flatter}), addressed dataset bias by proposing a flatter loss, improving performance across benchmarks. Similarly, Cao et al. (2020) tackled rank inconsistency in "Rank Consistent Ordinal Regression for Neural Networks with Application to Age Estimation" (CORAL-CNN \cite{cao2020rank}) using ordinal regression to enforce rank-monotonicity, boosting accuracy. Advanced loss function designs include Chen et al.'s "Delta Age AdaIN for Transformer-Based Age Estimation" (DAA \cite{chen2023daa}), which employs a Delta Age AdaIN operation with binary-coded reference groups, achieving significant improvements on UTKFace \cite{zhang2017utkface} and CelebA \cite{liu2015faceattributes}. These innovations demonstrate the critical role of loss functions in advancing age estimation.

\textbf{Feature Representation and Refinement: }
Refining feature representation has been another focal point for improving age estimation accuracy. Xia et al. (2020), in "Multi-Stage Feature Constraints Learning for Age Estimation" (MSFCL \cite{xu2021multistageage}), introduced multi-stage feature constraints during CNN training, enhancing model versatility. Zeng et al. (2020), in "Soft-Ranking Label Encoding for Robust Facial Age Estimation" (Soft-Ranking Label \cite{9145576}), proposed ordinal label encoding to provide nuanced supervision signals, improving robustness to noise and outliers. For forensic applications, Jeuland et al.'s "Assessment of Age Estimation Methods for Forensic Applications Using Non-Occluded and Synthetic Occluded Facial Images" (\cite{jeuland2022assessment}) explored feature refinement strategies under challenging conditions like synthetic occlusion. Bekhouche et al. (2024) extended this perspective with MSDNN \cite{bekhouche2024msdnn}, a multi-stage deep neural network that progressively captures hierarchical facial features while reusing information across stages, leading to improved efficiency and competitive results. 
These methods emphasize the importance of optimizing feature representation for reliable and adaptable models.

\textbf{Transformer-Based Approaches:  } The application of transformers to age estimation is an emerging research area with significant potential. Kuprashevich et al. (2023), in "MiVOLO"\cite{kuprashevich2023mivolo}, utilized a multi-input Vision Transformer (VOLO) architecture to process facial and full-body images, leveraging richer contextual cues for state-of-the-art results on UTKFace, Adience, and IMDB-W. Lixiong et al. (2023), in "SwinFace" \cite{lixiong2023swinface}, employed a Swin Transformer with Multi-Level Channel Attention (MLCA) modules, achieving a remarkable mean absolute error (MAE) on popular datasets. Chen et al. (2023) advanced the field with "Delta Age AdaIN" \cite{chen2023daa}, leveraging delta age information for nuanced estimation. Deng et al.'s work, "Joint Multi-Feature Learning for Facial Age Estimation" (JML \cite{deng2022joint}), demonstrated transformers' ability to model multi-feature interactions, extending their applicability to facial analysis tasks. Building on their earlier work with MiVOLO~\cite{kuprashevich2023mivolo}, Kuprashevich et al. (2025) introduced \emph{Beyond Specialization}~\cite{kuprashevich2024mivolo2}, where they assessed the capabilities of multimodal large language models (MLLMs) for age and gender estimation. This study highlighted both the potential and the limitations of general-purpose MLLMs compared to specialized deep models, providing an updated perspective on the evolving landscape of age estimation research.
These studies highlight transformers' transformative potential in age estimation through attention-based mechanisms.

\textbf{Holistic Approaches:}   Some methodologies adopt a holistic approach for broader facial analysis. Xie et al. (2019), in "Chronological Age Estimation under the Guidance of Age-Related Facial Attributes" (ChronoNN \cite{8662681}), simultaneously learned chronological age and related facial attributes, enhancing the understanding of aging patterns. Zhang et al. (2022), in "Cross-Dataset Learning for Age Estimation" (CDCNN \cite{zhang2022cross}), introduced a cross-dataset training framework that addresses dataset-specific inconsistencies, achieving state-of-the-art results on prominent benchmarks. In a complementary direction, relative age estimation has been introduced as an alternative paradigm, where the focus shifts from absolute regression to comparative prediction. The 2025 work on relative age estimation \cite{sandhaus2025relativeage} demonstrated how leveraging reference images with known ages allows models to refine predictions through differential regression.

These diverse methodologies, spanning hybrid models, innovative loss functions, refined feature representation, transformer-based approaches, and holistic strategies, collectively drive progress in face-based age estimation. Each cluster represents a significant step forward, providing valuable insights and tools for tackling the complexities of this challenging task.

%%%%%%%%%%%%%%%%%%%%%%%%%%%%%%%%%%%%%%%%%%%%%%%%%%%%%%%%%%%%%
 
%%%%%%%%%%%%%%%%%%%%%%%%%%%%%%%%%%%%%%%%%%%%%%%%%%%%%%%%%%%%%

\section{Proposed work}
%%%%%%%%%%%%%%%%%%%%%%%%%%%%%%%%%%%%%%%%%%%%%%%%
\label{sec:proposed_work}

\begin{figure*}
    \centering
    \includegraphics[width=1.1\linewidth]{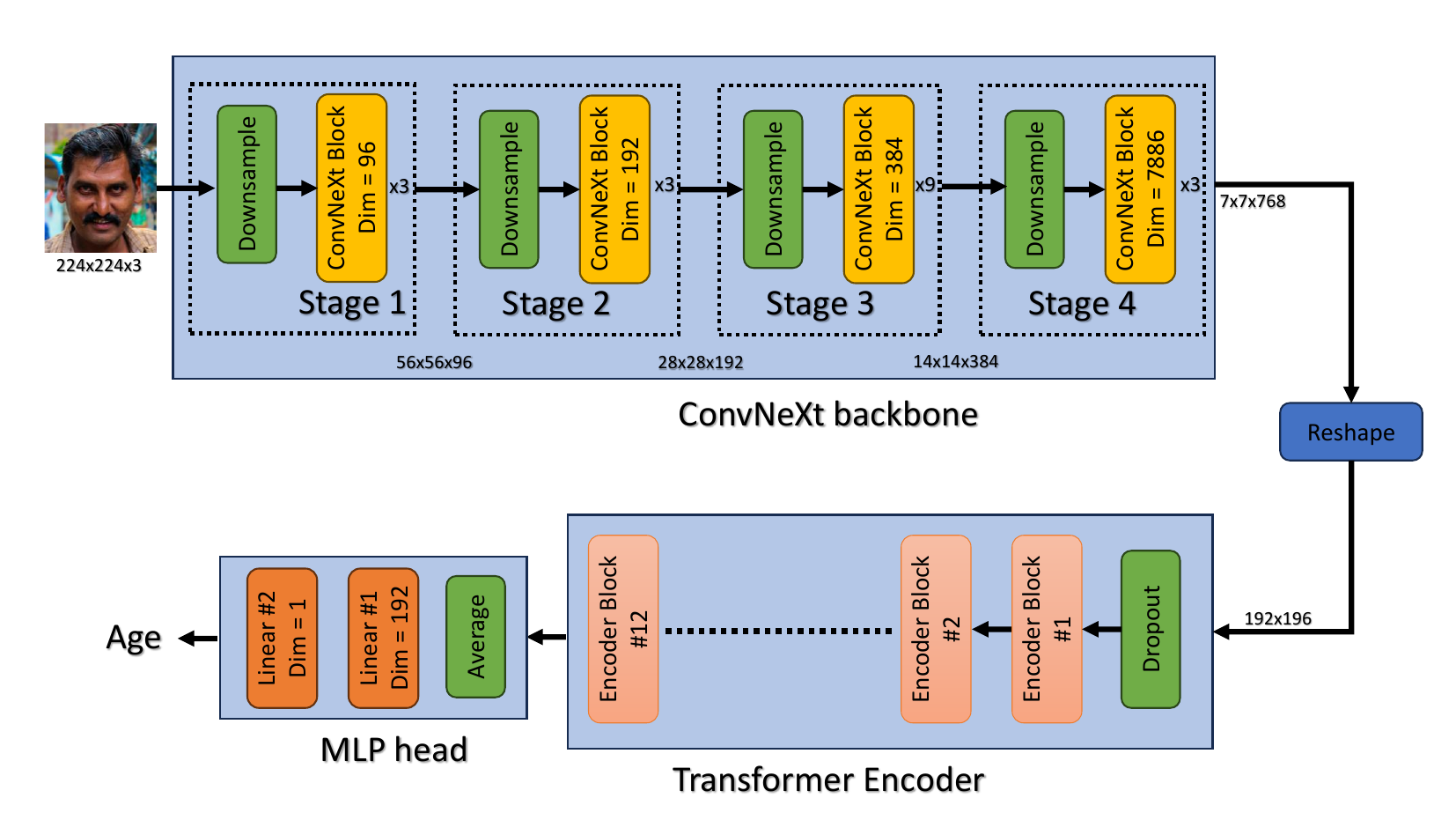}
    \caption{General structure of the proposed ConvNeXt-Transformer architecture.}
    \label{fig:general}
\end{figure*}

In this section, we present our innovative approach that leverages the capabilities of two advanced deep learning architectures: Transformers \cite{vaswani2017attention} and ConvNext \cite{liu2022convnet}. First, we give an overview of each architecture. Then, we explain the hybrid model based on the above two architectures, which improves the performance and robustness of age prediction from face images. The general architecture of our proposed model is shown in Figure \ref{fig:general}.

%-------------------------------------------------------------------------

\subsection{ConvNeXt}
\begin{figure}
    \centering
    \includegraphics[width=0.9\linewidth]{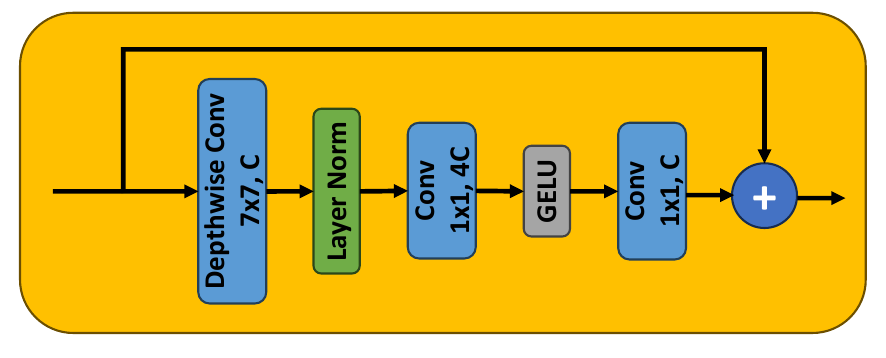}
    \caption{Details of a ConvNeXt Block.}
    \label{fig:convnext_block}
\end{figure}

ConvNeXt consists of several ConvNeXt blocks (see Figure \ref{fig:convnext_block}, each of which represents an advanced CNN architecture that proves to be a powerful solution for computer vision tasks \cite{liu2022convnet}. While ConvNeXt is inspired by the design principles of ViTs, it does not include the self-attention layers characteristic of ViTs. Instead, ConvNeXt modernizes the CNN structure by introducing transformative elements, effectively combining the strengths of CNNs with the advantages of transformer architectures. The ConvNeXt model uses a hierarchical arrangement of convolutional blocks that are incrementally extended to reflect certain aspects of the ViT design. However, it is important to emphasize that ConvNeXt retains the convolutional operations at its core and does not use the multi-head self-attention mechanisms found in transformers. The ConvNext block used is structured with a $7\times7$ depth-wise convolution, followed by LayerNorm, followed by a $1\times1$ depth-wise convolution with four times as many channels as the original convolution. This is followed by a GELU activation and finally a $1\times1$ depthwise convolution with the same number of channels as the first \cite{liu2022convnet}.

The authors of ConvNeXt argue that the architecture's success is due to its ability to learn both long-range and local dependencies between features. This allows ConvNeXt to learn a more comprehensive representation of the face, which leads to better age estimation performance.

As we will show in the experimental results section, ConvNext was able to achieve top results on the datasets used in this work (MORPH II, CACD, IMDB-Clean and AFAD), significantly outperforming previous methods. It is a good choice for tasks where accuracy and efficiency are important.

{\color{black} While ConvNeXt excels at capturing local spatial hierarchies and efficiently extracting features from facial images, it relies primarily on convolutional operations, which may limit its ability to capture global relationships and long-range dependencies within the image. Given that age-related cues can sometimes be distributed across different facial regions, a model capable of integrating both local and global information may further enhance age estimation accuracy. This motivates the integration of Vision Transformers (ViT), which are designed to model such global dependencies through self-attention mechanisms.}

\subsection{ViT}

{\color{black}Vision Transformers (ViT) bring a fundamentally different approach to image analysis by utilizing self-attention to model relationships between distant patches of an image. For age estimation, this ability allows the model to consider subtle and spatially dispersed age cues that may be overlooked by convolutional networks. By complementing ConvNeXt’s local feature extraction with ViT’s global context modeling, we aim to develop a more robust and accurate age estimation system.}

ViT is a type of transformer \cite{vaswani2017attention} for computer vision tasks. The basic structure of ViT, as outlined in the original paper, consists of a patch projection module in combination with a Transformer encoder similar to the architecture of BERT, but without the decoder component. Unlike traditional approaches that process images with CNNs, ViTs segment images into patches, convert each patch into a vector, and then reduce the dimensionality through a matrix multiplication step that represents the patch projection module \cite{dosovitskiy2020image}. These vector representations are then fed into a transformer encoder, which treats them as if they were token embeddings. Using this methodology, ViTs can effectively capture spatial relationships within images due to the use of self-attention mechanisms, making them a promising alternative to CNNs in computer vision applications.

\subsection{ConvNext-Transformer}
\label{sec:convnext_transformer}

The proposed model, as shown in Figure \ref{fig:general}, features a hybrid architecture combining the ConvNeXt backbone followed by a Transformer. This design aims to leverage the local feature extraction capabilities of ConvNeXt along with the global contextual understanding of the Transformer. Initially, the ConvNeXt backbone, organized into 4 stages, processes an input image of size $224 \times 224$. It produces features with a spatial resolution of $7 \times 7$ and a feature dimension of 768.  To adapt the feature space for the input into the Transformer, which excels with 2-dimensional sequence data, a comprehensive reshaping operation is executed. Initially, the spatial dimensions $7 \times 7$ of the feature maps produced by ConvNeXt are flattened, thus transforming the area into 49 consecutive feature vectors, while each vector retains a depth of 768 dimensions. Following this flattening, these 49 vectors are each subjected to a projection into a higher-dimensional space to match the required input structure of the Transformer. This elevation of dimensions is achieved through a learnable linear transformation, which maps each 768-dimensional vector to a new space meticulously crafted to fit the architectural needs of the Transformer, which is based on the ViT mention in the latter section, effectively creating an output with dimensions $192 \times 196$. Afterwards, the features, reshaped to dimensions of $192 \times 196$, are input into a Transformer that consists of 12 encoder blocks. Each block includes an attention module followed by a feedforward network. This arrangement allows the Transformer to efficiently process and integrate context from the input features. Finally, the output from the Transformer is directed to an MLP head. This module performs average pooling across the sequence dimension and includes one or two linear layers, depending on the configuration. For the age estimation task, a single linear layer output is fixed at a dimension of 1 to ensure consistency. In configurations with two linear layers, the dimensions of the output of the first layer and the input of the second can vary; in this research, we explored dimensions of 32, 64, 128, 192, and 256.

This hybrid design is expected to improve age estimation performance because it leverages the complementary strengths of ConvNeXt and Vision Transformers. ConvNeXt efficiently extracts local, fine-grained features such as wrinkles or skin texture, which are critical for assessing subtle age cues. However, age-related information can also be distributed across distant facial regions (e.g., hairline, jawline, and eyes), which ConvNeXt alone may not capture. By integrating a Vision Transformer branch, which models global relationships through self-attention, the hybrid architecture can aggregate both local and global features. This combination is particularly valuable in challenging scenarios where age indicators are subtle or spatially dispersed. As a result, the hybrid approach is expected to deliver higher accuracy and robustness than either backbone alone.

%-------------------------------------------------------------------------

\section{Experimental setup}

\subsection{Datasets}
\label{sec:data}

 We use three public face datasets.
 
\textbf{CACD (Cross-Age Celebrity Dataset \cite{chen2015face})}. CACD is a comprehensive dataset designed for age estimation and face recognition tasks. It contains a large collection of celebrity images spanning various ages and ethnicities. The dataset includes over 160,000 face images of more than 2,000 celebrities with age annotations, making it a valuable resource for age-related research.
% Reference: Chen, J., Song, L., Wang, L., Hu, X., & Yang, M. (2015). CACD: A benchmark database for continuous age estimation. In Proceedings of the 23rd ACM international conference on Multimedia (ACM MM).

The\textbf{ MORPH II \cite{Ricanek2006morph}} dataset is a widely used dataset for age progression and regression tasks. It comprises facial images of individuals of different ages, showing the aging process. MORPH II contains a substantial collection of more than 55,000 facial images with age labels, offering a diverse set of subjects and ages for age-related research.

\textbf{AFAD (Asian Face Age Dataset \cite{niu2016ordinal})} is a dataset specifically curated for age estimation tasks, focusing on Asian faces. It includes facial images of individuals from various Asian ethnicities, captured at different ages. The dataset contains over 160,000 images, providing a substantial resource for age-related research in the context of Asian populations.

The \textbf{IMDB-Clean dataset \cite{lin2021fpage}} is a refined benchmark for facial age estimation, created by filtering the original IMDB-WIKI dataset to remove images with noisy or unreliable labels. IMDB-Clean contains high-quality face images collected from the Internet Movie Database (IMDB), each annotated with an accurate age label. The dataset comprises more than 50,000 carefully curated images, representing a wide range of ages, ethnicities, and gender groups. This diversity, along with its improved label quality and standardized splits, makes IMDB-Clean a valuable resource for training and evaluating robust age estimation models in the presence of real-world variability.

The dataset splitting strategy varied across datasets depending on their structure and existing benchmarks. The same partitions of CORAL~\cite{cao2020rank} were applied on Morph II, CACD, and AFAD, while for IMDB-Clean, we followed its original split. In cases where identities appear across subsets (as in CACD and Morph II), the random splitting reflects the established methodology in previous works and aims to preserve age and demographic balance across all phases. A more detailed justification of this choice is provided in \textbf{Section 4.3 Experimental Setup}.

\subsection{Preprocessing}
\label{sec:preprocessing}
Preprocessing steps play a crucial role in preparing the data for age estimation. They ensure that the data are structured, diverse, and appropriately scaled, enabling the model to learn effectively and make accurate predictions when faced with real-world challenges. Proper data preparation and augmentation are key factors in enhancing the model's performance and its ability to generalize across different age estimation scenarios.

Once the faces are detected, all faces are resized to $224\times224$ pixels. Resizing the faces to a fixed size is crucial to ensure that all input images have the same dimensions, which is required by the used models.

\subsection{Experimental Setup}
\label{sec:parameters}
Firstly, configuration parameters such as the number of epochs, batch size, loss function, and learning rate are established. The training and validation data are then loaded and organized using data loaders. After training the model, it is tested on a separate test set, and the results are stored for further analysis. The AdamW optimizer is used for training, and a learning rate scheduler is configured according to the chosen strategy, which may include reducing the learning rate on a plateau or following a warmup cosine schedule.

In our experiments, we used an adaptive regression loss function proposed by \cite{feng2019adaptive}, defined as:
\[
L(y, \hat{y}) = \frac{1 + \sigma}{N} \sum_{i=1}^{N} \frac{(y_i - \hat{y}_i)^2}{|y_i - \hat{y}_i| + \sigma}
\]
where \( y_i \) is the ground-truth age, \( \hat{y}_i \) is the predicted age, \( N \) is the batch size, and \( \sigma \) is a positive constant controlling the adaptiveness (set to 2 in our case). This loss penalizes small errors more strongly while reducing the influence of large errors, making it a suitable balance between MAE and MSE for age estimation. Its error-aware weighting improves robustness to outliers while maintaining precision for near-correct predictions.

This setup covers the architecture, data processing, training, and evaluation for a model that can integrate ConvNeXt and ViT architectures, offering flexibility in selecting different models and training strategies. To ensure comparability, preprocessing, data preparation, and model architecture are based on the same features, allowing for consistent fine-tuning across various datasets under study.

In our age estimation research, we opted for the method of random splitting when dividing the albums datasets, in a way, a person's images in different age stances might appear in both sets. This choice was deliberate, aimed at achieving robust and accurate results. By doing so, we ensured that our model's training, validation, and testing phases received an equitable representation of the dataset, promoting its adaptability to real-world scenarios. The use of random splitting allowed us to conduct fair evaluations, guarding against any inadvertent data biases and mitigating the risk of overfitting. This approach adheres to widely used protocols in prior works, including DEX~\cite{rothe2018deep}, CORAL~\cite{cao2020rank}, and MiVOLO~\cite{kuprashevich2023mivolo}, which all use random image-level splits on Morph II and CACD without enforcing subject exclusivity.
While subject-exclusive splitting may offer a better assessment of identity generalization, it is not commonly enforced in these benchmarks. We acknowledge its value and consider it an important direction for future work.
% Moreover, maintaining a balanced age distribution across subsets through random splitting bolstered our model's ability to predict ages across diverse demographics. Our decision adheres to established best practices and reinforces the reliability and precision of our age estimation approach

Our data preprocessing pipeline leverages several augmentation techniques to improve our model's robustness and its capacity to handle diverse images effectively. Of note, the "Random Square Erasing" technique is a valuable addition, simulating partial occlusions in the images. This helps our model learn to focus on key facial features and disregard irrelevant information, making it adept at age estimation even when portions of a face are obscured. Additionally, our pipeline includes other augmentations such as random cropping, horizontal flipping, color jittering, random rotation, and Gaussian blur, providing an array of strategies to prepare the data for training. All the previous steps led us to be able to fine-tune on the famous MORPH II album, CACD,  AFAD and the recently introduced IMDB-Clean datasets.

For the IMDB-Clean dataset, we additionally applied the MiVOLO \cite{kuprashevich2023mivolo} face-cropping protocol, ensuring that only the facial region was used for training and evaluation. This decision was made to reduce the impact of background and non-face information present in the IMDB-Clean images, thus focusing the model on the most age-informative features.

In all experiments, the architectures evaluated include ConvNeXt, ViT, and their hybrid combination. We systematically explored a range of configurations, including fully connected head sizes (32, 64, 128, 192, 256), and multiple loss functions (MAE, MSE, Huber, Adaptive, WeightedMSE). Model training uses the AdamW optimizer, with batch sizes between 64 and 256 and learning rates from 1e-2 to 1.5e-5, depending on the experimental run. Learning rate scheduling strategies compared include ReduceLROnPlateau, CosineAnnealingLR, WarmupCosineSchedule, and OneCycleLR. All experiments are conducted for up to 500 epochs, with early stopping based on validation MAE. Both training from scratch and transfer learning are considered, with ConvNeXt initialized from the best prior checkpoint and ViT from ImageNet-pretrained weights when appropriate.

\subsection{Two-stage training}
\label{sec:decision_layer}
Our age estimation models are the result of a two-phase learning process. In the initial phase, we utilize the extensive ImageNet dataset for pretraining. This phase equips our models with a foundational understanding of visual features, textures, and shapes. It acts as a critical stepping stone in their journey towards age estimation expertise. ImageNet's vast and diverse collection of images across numerous categories serves as a valuable source of visual knowledge. Pretraining on ImageNet is akin to providing our models with a "worldly education" in the realm of visual data.

Following this pretraining phase, our models are fine-tuned on the CACD, AFAD, IMDB-Clean and MORPH II datasets. These datasets are specifically tailored for age estimation tasks. Fine-tuning refines the knowledge acquired during pretraining, making it age-specific. The combination of these datasets introduces our models to the intricacies of age-related features, helping them become age estimation experts. This two-phase process, grounded in pretraining on ImageNet and fine-tuning on age-specific datasets, forms the backbone of our age estimation approach. It leverages both generic and specialized knowledge, resulting in models that excel in estimating age from facial images. The following sections of our article delve deeper into the specific choices made for data preprocessing, model architectures, and training strategies, highlighting the intricate details of our approach.

%-------------------------------------------------------------------------
\section{Ablation Study and Experimental Results}
\label{sec:experiments}

The proposed method is evaluated on several benchmark datasets and the results show that it outperforms existing state-of-the-art methods in terms of age estimation accuracy.

% \subsection{Ablation Study Results with Explanation}
% \label{sec:ablation_study}

In our detailed ablation study, we investigate the effects of modifying various hyperparameters on the performance of the proposed ConvNeXt model, then the Vision Transformer model, which leads us to the combination of both in the final ConvNeXt-Transformer model, with a specific focus on quantifying these effects in terms of MAE. The experiments are conducted across three diverse datasets: MORPH II, CACD, IMDB-Clean and AFAD

MAE is one of the most widely used evaluation metrics in age estimation. It measures the average absolute difference between the predicted ages $\hat{y}_i$ and the ground truth ages $y_i$ over $N$ samples:
\begin{equation}
\text{MAE} = \frac{1}{N} \sum_{i=1}^{N} \left| y_i - \hat{y}_i \right|.
\end{equation}
A lower MAE indicates better accuracy of the model.

The tables presented show numerous experiments, each featuring unique hyperparameter configurations. Subject-Independent data splitting is consistently employed, indicating its effectiveness for generalization. The AdamW optimization algorithm is used consistently, providing stable and effective training. We start our journey by experiencing different ways to find our optimal ConvNeXt model, and then later, we transition to the Transformer model, mirroring the investigative process to identify the optimal model too. 

This iterative path ultimately guides us in crafting the final model, we combine the two best-performing fine-tuned Transformer-ConvNeXt models to make our final model.

\subsection{\textbf{ConvNeXt model alone}}
\label{sec:ablation_study_convnext}

In the following experiments, we fine-tuned a pre-trained ConvNeXt model, originally trained on ImageNet, by modifying the output head to assess the effect of different configurations on facial age estimation accuracy. Specifically, we replaced the last linear layer with either one or two linear layers and evaluated their impact on MAE across three datasets: MORPH II, CACD, and AFAD. Table \ref{tab:nb_fc_convnext_table} summarizes the results, showing that using two linear layers in the output head consistently yielded lower MAE compared to a single linear layer for all datasets. For the MORPH II dataset, adding a second linear layer reduced the MAE from 2.44 to 2.29, indicating a notable improvement in accuracy, likely due to the added representational capacity of the second layer enabling the model to better capture the data distribution. In the CACD dataset, the MAE decreased from 4.42 to 4.40 with two linear layers, demonstrating a consistent improvement, although less pronounced than in MORPH II. Similarly, for the AFAD dataset, the MAE is almost the same. These findings suggest that increasing the number of linear layers in the output head positively impacts MAE by enhancing the model's ability to generalize across different datasets.

\begin{table}[h]
    \caption{Effect of Number of Linear Layers in the Output Head}
    \label{tab:nb_fc_convnext_table}
    \centering
    \begin{tabular}{|l|c|c|}
        \hline
        Dataset & Linear Layers & MAE \\
        \hline
        \textbf{MORPH II} & 2 & \textbf{2.29} \\
        MORPH II & 1 & 2.44 \\
        \textbf{CACD} & 2 & \textbf{4.40} \\
        CACD & 1 & 4.42 \\
        \textbf{AFAD} & 2 & \textbf{3.12} \\
        \textbf{AFAD} & 1 & \textbf{3.12} \\
        \hline 
    \end{tabular} 
\end{table}

To further optimize the output head design, we explored the impact of the size of the last linear layer on the model's performance, experimenting with various configurations to determine the optimal layer size for training. Table \ref{tab:size_fc_convnext_table} illustrates that, for the MORPH II dataset with two linear layers in the output head, a size of 256 for the last linear layer yields the lowest MAE of 2.29. Reducing the size of this layer to 192 slightly increased the MAE to 2.40, and further reductions in size to 128, 64, and 32 resulted in progressively higher MAE values of 2.41, 2.44, and 2.42, respectively. These results show that larger linear layer sizes help the model estimate facial age more accurately. The improvement is likely because the larger layers have more capacity, allowing the model to better understand and represent the details of the dataset. However, using very large linear layers can also increase computing time and may require more training to achieve the best performance.

\begin{table}[h]
    \caption{Effect of Second Linear Layer Size}
    \label{tab:size_fc_convnext_table}
    \centering
    \begin{tabular}{|l|c|c|}
        \hline
        % Dataset & 
        Layer Size & MAE \\
        \hline
        % MORPH II & 
        256 & \textbf{2.29} \\
        192 & 2.40 \\
        128 & 2.41 \\
        64 & 2.44 \\
        32 & 2.42 \\

        \hline
    \end{tabular}
\end{table}

From the results shown in Table \ref{tab:nb_epochs_convnext_table}, it is clear that 500 epochs produce the best results. However, after 100 epochs, there is little improvement in MAE, suggesting that further training beyond this point offers limited additional benefit.

\begin{table}[h]
    \caption{Effect of Number of Epochs}
    \label{tab:nb_epochs_convnext_table}
    \centering
    \begin{tabular}{|l|c|c|}
        \hline
        Epochs & MAE \\
        \hline
        100 & 2.36 \\
        500 & 2.29 \\
        \hline
    \end{tabular}
\end{table}

\textbf{Overall Conclusion for the ConvNeXt model:}
In summary, our comprehensive ablation study conducted on the MORPH II, CACD, and AFAD datasets clarify the influence of different model configuration factors on MAE. These insights underscore the significance of customizing ConvNeXt model architectures based on dataset characteristics, thereby guiding future model tuning and parameter selection to enhance performance across diverse albums.

The most optimal model identified in this study, which attained the lowest MAE, and which will qualify to be used in the final model, was trained on the MORPH II dataset using two linear layers, with the batch size of the last linear layer set to 256 and trained for 500 epochs.

\subsection{\textbf{Transformer model (ViT) alone}}
\label{sec:ablation_study_vit}

To begin, we followed the same approach used for the ConvNeXt model (explained in the previous section \nameref{sec:ablation_study_convnext}), but applied it to the ViT model. This approach led us to use 500 epochs and two linear layers for optimal performance. The best results achieved in the MORPH II dataset using pretrained ViT reached an MAE of 2.453. In the subsequent phase, we investigated the influence of different sizes for the last linear layer on the MORPH II dataset. Our exploration revealed that a size of 192 produced the best outcome, resulting in an MAE of 2.4 (see Table \ref{tab:size_fc_vit_table}).

\begin{table}[h]
    \centering
    \caption{Impact of Size of linear layer Batch on MAE for MORPH II}
    \label{tab:size_fc_vit_table}
    \begin{tabular}{|c|c|}
        \hline
        Linear layer size & MAE \\
        \hline
        256 & \textbf{2.47} \\
        128 & 2.50 \\
        32 & 2.55 \\
        \hline
    \end{tabular}
\end{table}

To improve performance, we introduced a new learning rate scheduler that led to better results (see Table \ref{tab:scheduler_vit_table}). Specifically, we switched from the manual scheduler to the warming cosine scheduler. The Warmup Cosine Scheduler starts with a lower learning rate that gradually increases over a set number of steps, before decreasing in a cosine pattern. This approach helps the model stabilize early in training, leading to better overall performance. In contrast, the Manual Scheduler uses a fixed learning rate throughout the training. Our experiments showed that the ViT model performed significantly better on the MORPH II and CACD datasets with the Warmup Cosine Scheduler, although there is still room for improvement on the AFAD dataset.

\begin{table}[h]
    \centering
    \caption{Impact of scheduler on MAE for MORPH II}
    \label{tab:scheduler_vit_table}        
    \begin{tabular}{|l|l|c|}
        \hline
        Scheduler & MAE \\
        \hline
        Warmup Cosine Schedule &\textbf{2.47} \\
        Manual Scheduler & 4.39 \\
        \hline
    \end{tabular}
\end{table}

\textbf{Overall Conclusion for ViT Model:}
Our study showed that the ViT model performed better than the current best methods SOTA on the MORPH II and AFAD datasets. For the MORPH II dataset, the best result was an MAE of 2.47, achieved with a linear layer size of 256 (see Table \ref{tab:size_fc_vit_table}) and the Warmup Cosine Schedule for learning rate adjustment (see Table \ref{tab:scheduler_vit_table}). However, the results on the CACD dataset indicated that there is still room for improvement. The best model from this study, which used a linear layer size of 256 and the Warmup Cosine Schedule, achieved the lowest MAE on each dataset. This model will be used in the experiments of the hybrid model, aiming to combine the strengths of ViT and improve its performance further.

\subsection*{\color{black}{5.3 Training Loss Curves}}
\label{sec:train_loss_curves_study}

\begin{figure}
    \centering
    \Huge
    \includegraphics[width=1.\linewidth]{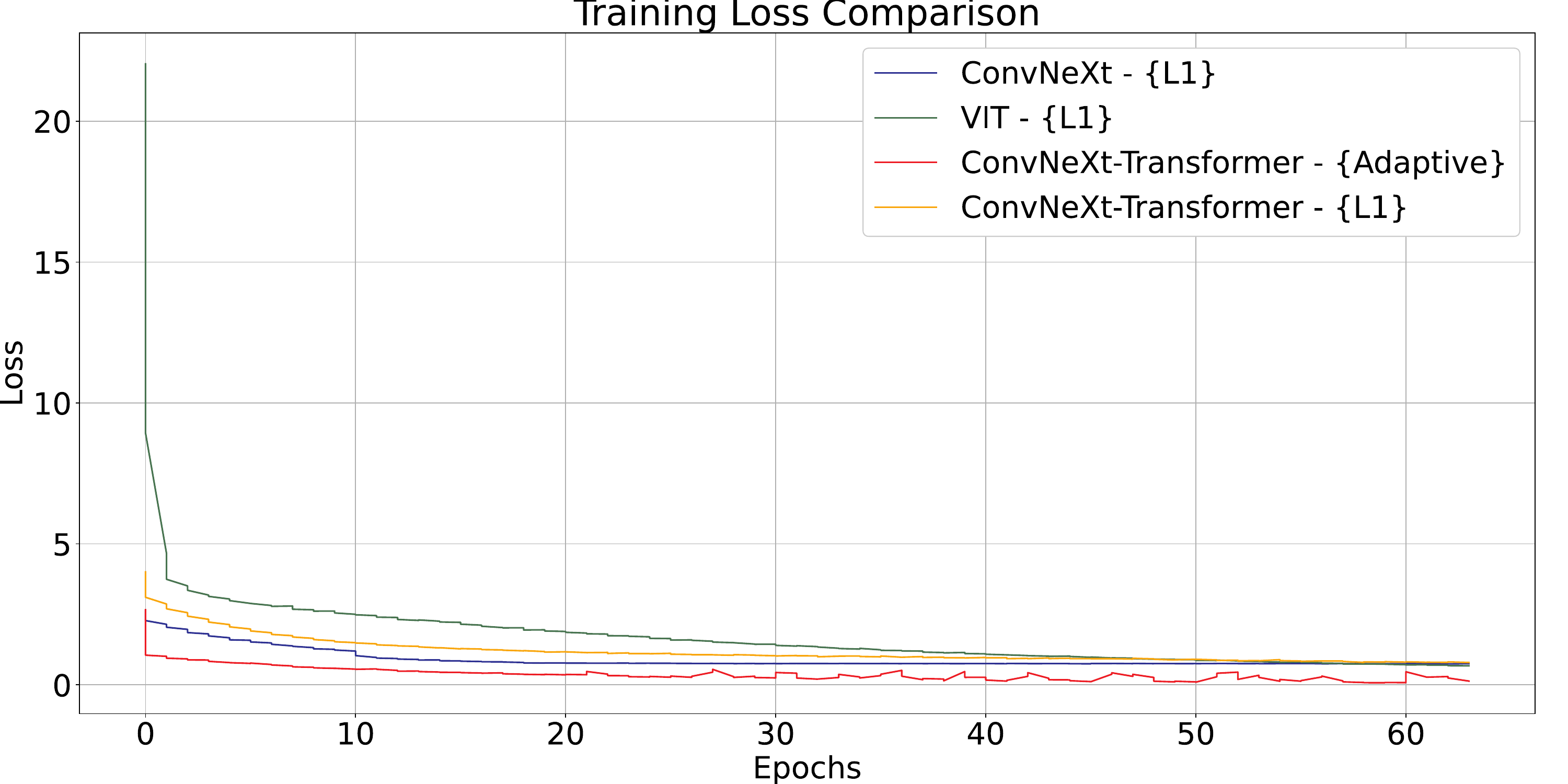}
    \caption{Training Loss vs. Epochs for Different Models. The plot shows the training loss values plotted against the number of epochs for three different models on the MORPH II album dataset:  ConvNeXt with the MAE loss function, ViT with the MAE loss function, ConvNeXt-Transformers with both, the MAE loss function and the adaptive one.}
    \label{fig:training_loss_comparison}
\end{figure}

The training loss curves depicted in Figure \ref{fig:training_loss_comparison} illustrate the convergence behavior of three models over the course of training in the MORPH II album dataset. We observe that ConvNeXt, ViT and ConvNeXt-Transformer with the MAE loss function model exhibit a gradual decrease in loss with epochs, indicative of effective learning. In contrast, when using the Adaptive loss function on the ConvNeXt-Transformer model, it shows fluctuations in loss, particularly during the late epochs, suggesting potential challenges in convergence with these architectures. In general, these results provide information on the training dynamics of the models and underscore the importance of model selection and optimization strategies.

\setcounter{subsection}{3}

\subsection{\textbf{Hybrid model}}
\label{sec:hybrid_model}

The observed improvement in performance, particularly in CS@5 and MAE metrics, can be directly attributed to the hybrid model’s ability to combine fine-grained local features from ConvNeXt with the global context captured by the Vision Transformer. As described in section \textbf{\ref{sec:convnext_transformer} ConvNext-Transformer}, this synergy enables the model to handle both concentrated and dispersed age cues, which explains the boost in robustness and accuracy observed across all datasets, especially on the challenging IMDB-Clean benchmark. The ablation results further confirm that the integration of both components outperforms either model alone, validating our design choices.

Based on the results of our ablation study, the following configuration was adopted for our final ConvNeXt-Transformer model training. Images were resized to 224×224 pixels. For IMDB-Clean, we used MiVOLO face-cropping, a batch size of 96, an initial learning rate of 1.5e-5, the WeightedMSE loss, OneCycleLR scheduling, and trained for up to 500 epochs with early stopping (patience 3). For the other datasets (MORPH II, CACD, AFAD), we used a batch size of 128, only trained for 100 epochs which looked enough, applied the Adaptive loss, and used Warmup Cosine Scheduler. In all cases, images were resized to 224×224, standard augmentations were applied, and the model’s fully connected head had 256 units. ConvNeXt weights were initialized from our best prior ImageNet checkpoint, and ViT from ImageNet-pretrained weights. Trainings were conducted using the Nvidia RTX 3090 GPUs with \textit{float32} precision.

Our ablation study identified the two optimal models to be used in our final hybrid model. This final model, a combination of ConvNeXt and Transformer, was compared with multiple state-of-the-art (SOTA) methods using the MORPH II dataset, with the results reported in Table \ref{tab:sota_morph2}. The results of previous methods are sourced from their respective publications, adhering to the same protocol specifications as ours. The MAE values demonstrate that our final model performs admirably, achieving strong results also on the CACD (Table \ref{tab:sota_cacd}), and AFAD (Table \ref{tab:sota_afad}) datasets.

\begin{table}[h!]
\caption{Our study evaluates age estimation performance using the test subset of} the Morph II dataset. We compare our results with previous SOTA schemes
\label{tab:sota_morph2}
\centering
\begin{tabular}{|l|c|c|}

\hline
Approach & Year & MAE  \\ \hline
CE        \cite{rothe2018deep} & 2018 & 6.54  \\ 
AGEn \cite{8141981}        & 2018 & 6.40  \\ 
DLDL-v2 \cite{gao2018age}        & 2018 & 5.80  \\ 
SGD  \cite{akbari2021flatter}   & 2020 & 5.69  \\ 
GJM  \cite{akbari2021does}        & 2021 & 5.63  \\ 
CasCNN \cite{wan2018auxiliary} & 2018 & 3.30  \\ 
OR-CNN \cite{niu2016ordinal}        & 2016 & 3.27  \\ 
Multi-Task \cite{8007264}  & 2020 & 3.00   \\ 
Google-Net \cite{Dagher2021Facial}  & 2021 & 2.94    \\ 
DRFss  \cite{shen2018deep}  & 2018 & 2.91  \\ 
SRL  \cite{9145576} & 2020 & 2.83  \\ 
CE-MV \cite{8578652}  & 2018 & 2.80  \\ 
deep-JREAE \cite{Tian2021Facial} & 2021 & 2.77 \\ 
CDCNN \cite{zhang2022cross}        & 2022 & 2.76  \\ 
Adaptive-CNN \cite{Dornaika2020robust}  & 2020 & 2.75  \\ 
SADAL  \cite{liu2020similarity}  & 2020 & 2.75  \\ 
MSFCL \cite{xia2020multi}        & 2020 & 2.73  \\ 
Wasserstein \cite{9457089}        & 2021 & 2.71  \\ 
CNN \cite{8662681}        & 2019 & 2.69  \\ 
DEX \cite{rothe2018deep}        & 2018 & 2.68  \\ 
CORAL-CNN \cite{cao2020rank}        & 2020 & 2.64  \\ 
ADPF \cite{wang2022improving}        & 2022 & 2.71  \\ 
Multi-Stage DNN \cite{bekhouche2024msdnn}   & 2024 & 2.59 \\ 
FO \cite{jeuland2022assessment}   & 2022 & 2.53 \\ 
RelativeAge \cite{sandhaus2025relativeage}   & 2025 & 2.47 \\ 
JML \cite{deng2022joint} & 2022 & 2.42  \\ \hline \hline
ConvNeXt    & Ours & 2.29 \\ 
ViT    & Ours & 2.47 \\ 
\textbf{ConvNeXt-Transformer}    &  \textbf{Ours} & \textbf{2.26} \\ \hline 
\end{tabular} 
\end{table}

\begin{table}[h!]
\caption{Our study evaluates age estimation performance using the test subset of the CACD dataset. We compare our results with previous SOTA schemes}
\label{tab:sota_cacd}
\centering
\begin{tabular}{|l|c|c|}
\hline
Model &  Year & MAE \\ \hline
CORAL         \cite{cao2020rank} & 2020 &  5.35    \\ 
RelativeAge \cite{sandhaus2025relativeage}   & 2025 & 5.27 \\ 
Multi-Stage DNN \cite{bekhouche2024msdnn}   & 2024 & 4.90 \\ 
DRFs         \cite{shen2018deep} & 2018 &  4.68    \\ 
RNDF         \cite{rndf2021}     & 2021 &  4.60      \\ 
MWR          \cite{shin2022moving}\textsuperscript{\dag} & 2022 &  4.41     \\ 
\hline \hline
ConvNeXt    & Ours & 4.40 \\ 
ViT    & Ours & 4.71 \\ 
\textbf{ConvNeXt-Transformer}    & \textbf{Ours} & \textbf{4.35} \\ \hline
\end{tabular} 
\end{table}

\textsuperscript{\dag}MWR result is reported on the CACD training subset, not the test set.

\begin{table}[h!]
\caption{Our study evaluates age estimation performance using the test subset of} the AFAD dataset. We compare our results with previous SOTA schemes
\label{tab:sota_afad}
\centering
\begin{tabular}{|l|c|c|}
\hline
Model &  Year & MAE \\ \hline
SimLoss \cite{simloss2020kobs}        & 2016 & 3.95  \\ 
OR-CNN \cite{niu2016ordinal}        & 2016 & 3.51  \\ 
CORAL\cite{cao2020rank}             &  2020 & 3.48  \\
Multi-Stage DNN \cite{bekhouche2024msdnn}   & 2024 & 3.25 \\ 
\hline \hline
ConvNeXt    & Ours & 3.12 \\ 
ViT    & Ours & 3.43 \\ 
\textbf{ConvNeXt-Transformer}    & \textbf{Ours} & \textbf{3.09} \\ \hline
\end{tabular}
\end{table}

\begin{table}[h!]
\caption{Our study evaluates age estimation performance using the test subset of the IMDB-Clean dataset. We compare our results with previous SOTA schemes. The best results are in bold.}
\label{tab:sota_imdb_mae}
\centering
\begin{tabular}{|l|c|c|c|}

\hline
Approach & Year & MAE)  \\ \hline
FP-Age \cite{lin2021fpage}        & 2023 & 4.68  \\  
GLA-Age \cite{LIU2025glaage}   & 2025 & 4.55 \\ 
VOLO-D1 (face only) \cite{kuprashevich2023mivolo}        & 2023 & 4.29  \\ 
% MCGRL \cite{shou2024mcgrl}}        & 2024 & 80.1  \\  
\hline \hline
ConvNeXt    & Ours & 4.23 \\ 
ViT    & Ours & 5.1  \\ 
\textbf{ConvNeXt-Transformer}    &  \textbf{Ours} & \textbf{4.22} \\\hline 
\end{tabular} 

\end{table}

\subsection{\textbf{Cumulative Distribution of Absolute Error}}
\label{sec:cdae_study}
 
The CS metric evaluates the percentage of age estimations that fall within a given error margin $k$ years from the ground-truth.  
It is defined as:
\begin{equation}
\text{CS}(k) = \frac{\#\{\,i : |y_i - \hat{y}_i| \leq k\,\}}{N} \times 100\%,
\end{equation}
where $\#\{\,i : |y_i - \hat{y}_i| \leq k\,\}$ counts the number of samples for which the absolute error is less than or equal to $k$.  
For example, CS@5 gives the proportion of predictions that are within $5$ years of the true age.  
Higher CS values indicate better performance.

\begin{figure}[h!]
    \centering
    \Huge
    \includegraphics[width=1.05\linewidth]{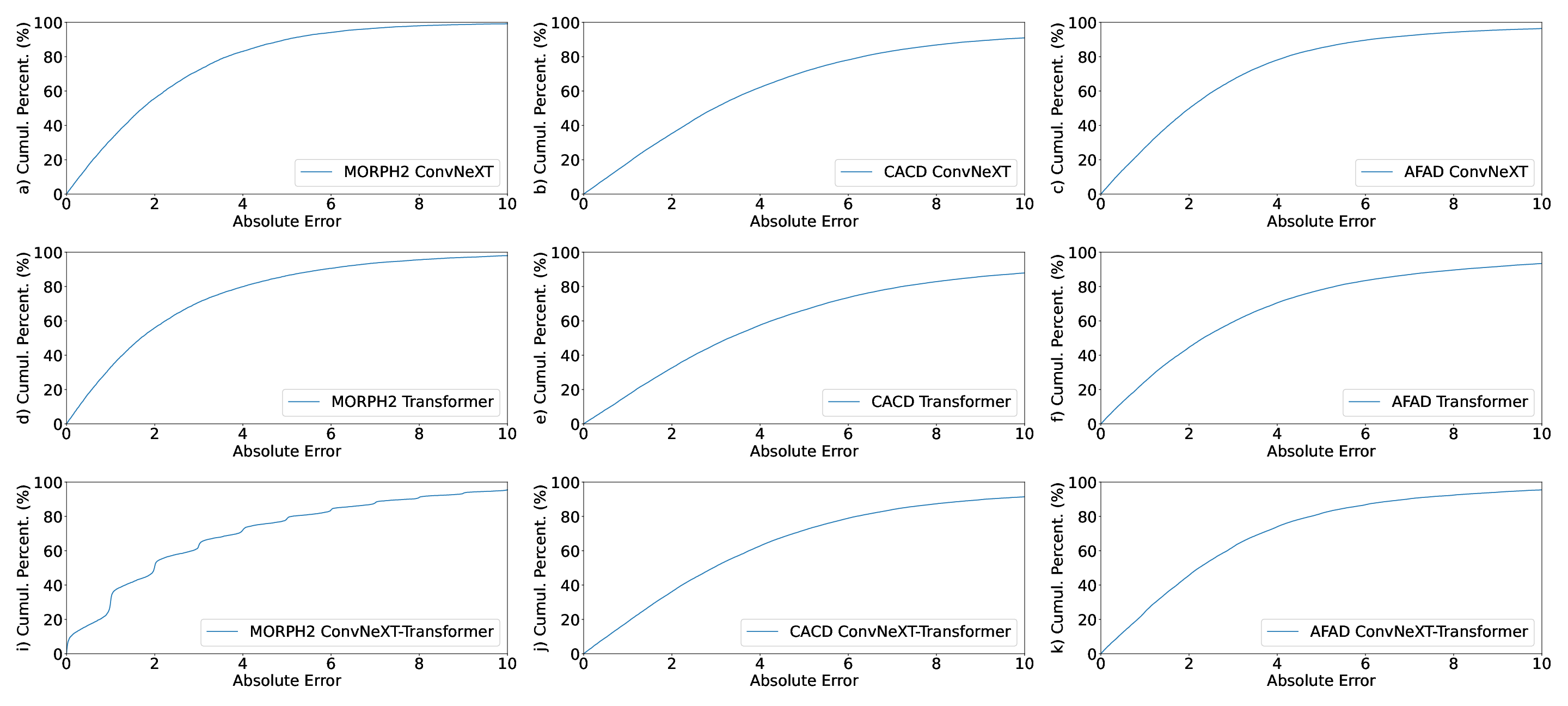}
    \caption{Cumulative Distribution of Absolute Error for 3 Different Models on the 3 Different Albums (MOPRH2, CACD and AFAD). The cumulative distribution plots provide an insightful comparison of model performance, showcasing the spread and density of absolute errors across the dataset. The example shown here demonstrates the best degrees of accuracy and precision we achieved in each model, offering valuable insights into their respective capabilities on the different dataset we're working on.}
    \label{fig:cumul_dist_mae}
\end{figure}

The Cumulative Distribtion of MAE curves for our three models are depicted in Figure \ref{fig:cumul_dist_mae}, which includes the ConvNeXt model (a, b \& c), the ViT model (d, e \& f) and  the Hybrid model (g,h \& i), which were evaluated across MORPH II, CACD, and AFAD datasets respectively. The results indicate that ConvNeXt models consistently achieve higher accuracy, especially on MORPH II, where they produced the best area under curve (AUC). Across all models on the different datasets, at least 90\% of predictions had an absolute error margin under 10 years, demonstrating the robustness and reliability of our models. These findings suggest that ConvNeXt, ViT and our ConvNeXt-Transformer models are particularly well-suited for age estimation tasks, achieving remarkable performance and providing a reliable approach for future applications in this domain.\\

To provide a more direct comparison with the literature, we also report the cumulative score at a 5-year error margin (CS@5) in Tables ~\ref{tab:sota_morph2_cs5}, ~\ref{tab:sota_cacd_cs5} and ~\ref{tab:sota_imdb_cs5}. Our hybrid ConvNeXt-Transformer models achieve CS@5 scores that are competitive with, or surpass, many previously published methods. For example, on IMDB-Clean (\autoref{tab:sota_imdb_cs5}), our model achieves a CS@5 of 67.71\%, outperforming established approaches like FP-Age and VOLO-D1, while on CACD (\autoref{tab:sota_cacd_cs5}), our ConvNeXt-Transformer reaches 72\% just like DRF and exceeding DEX. These results highlight that our models are not only robust across the entire error range, as shown by the cumulative error curves, but also deliver strong accuracy under the commonly used CS@5 metric. Overall, our models provide a reliable and state-of-the-art solution for age estimation across diverse benchmarks.\\

\begin{table}[h]
\caption{Our study evaluates age estimation performance using the Morph II dataset. We compare our results with previous SOTA schemes. We count cumulative scores for age errors within a 5-year range.The best results are in bold.}
\label{tab:sota_morph2_cs5}
\centering
\begin{tabular}{|l|c|c|}

\hline
Approach & Year & CS@5(\%)  \\ \hline
OR-CNN \cite{niu2016ordinal}        & 2016 & 74.3  \\ 
DEX \cite{rothe2018deep}        & 2018 & 84.9  \\  
Multi-Stage DNN \cite{bekhouche2024msdnn}   & 2024 & 86.66 \\ 
\hline \hline
\textbf{ConvNeXt}    & \textbf{Ours} & \textbf{90.1} \\ 
ViT    & Ours & 86.33 \\ 
ConvNeXt-Transformer    &  Ours & 78.65 \\\hline 
\end{tabular}

\end{table}

\begin{table}[h]
\caption{Our study evaluates age estimation performance using the CACD dataset. We compare our results with previous SOTA schemes. We count cumulative scores for age errors within a 5-year range.The best results are in bold.}
\label{tab:sota_cacd_cs5}
\centering
\begin{tabular}{|l|c|c|}

\hline
Approach & Year & CS@5(\%)  \\ \hline
DEX \cite{rothe2018deep}        & 2018 & 68.2  \\  
\textbf{DRF} \cite{shen2018deep}        & \textbf{2018} & \textbf{72.8}  \\ 
Multi-Stage DNN \cite{bekhouche2024msdnn}   & 2024 & 65.31 \\ 
\hline \hline
ConvNeXt    & Ours & 71.2 \\ 
ViT    & Ours & 66.3 \\ 
\textbf{ConvNeXt-Transformer}    &  \textbf{Ours} & \textbf{72.0} \\\hline 
\end{tabular} 

\end{table}

\begin{table}[h]
\caption{Our study evaluates age estimation performance using the IMDB-Clean dataset. We compare our results with previous SOTA schemes. We count cumulative scores for age errors within a 5-year range.The best results are in bold.}
\label{tab:sota_imdb_cs5}
\centering
\begin{tabular}{|l|c|c|c|}
\hline
Approach & Year  & CS@5(\%)  \\ \hline
FP-Age \cite{lin2021fpage}        & 2023  & 63.78  \\  
GLA-Age \cite{LIU2025glaage}   & 2025 & 65.06 \\ 
VOLO-D1 face \cite{kuprashevich2023mivolo}        & 2023  & 67.68  \\ 
\hline \hline
ConvNeXt    & Ours & 58.56 \\ 
ViT    & Ours & 55.07 \\ 
\textbf{ConvNeXt-Transformer}    &  \textbf{Ours} & \textbf{67.71} \\\hline 
\end{tabular} 

\end{table}

\break

%-------------------------------------------------------------------------

\section{Conclusions and future work}
\label{sec:conclusion}

In this paper, we demonstrate the remarkable potential of combining ConvNeXt—an advanced Convolutional Neural Network (CNN) architecture described by Liu et al.                     \cite{liu2022convnet} as "a CNN for the 2020s"—with the Transformer model. This innovative synergy leverages the strengths of both architectures, enabling us to capture local and global facial information more effectively. As a result, our approach significantly enhances the regression accuracy of age estimation models.

Our hybrid approach achieved state-of-the-art performance on age estimation tasks across multiple well-known facial datasets. These results validate the method's ability to extract salient features and effectively manage variability among faces, highlighting the complementary strengths of advanced CNNs and Transformer models.

Looking ahead, this work represents a foundational step toward extending such hybrid architectures to tackle other challenging supervised learning tasks, such as classification and segmentation. With appropriate adaptations, the proposed model shows promise for broader applications, further bridging the gap between deep learning paradigms and practical challenges in computer vision. By focusing on this convergence of techniques, we aim to pave the way for continued exploration and advancements in the field.

\end{document}